\documentclass{article}
\usepackage{arxiv}
\usepackage[utf8]{inputenc} % allow utf-8 input
\usepackage[T1]{fontenc}    % use 8-bit T1 fonts
\usepackage{hyperref}       % hyperlinks
\usepackage{url}            % simple URL typesetting
\usepackage{booktabs}       % professional-quality tables
\usepackage{amsfonts}       % blackboard math symbols
\usepackage{nicefrac}       % compact symbols for 1/2, etc.
\usepackage{microtype}      % microtypography
\usepackage{lipsum}
\usepackage{graphicx}
\usepackage{amsmath}
\usepackage{amssymb}
\usepackage{caption, floatrow}
\usepackage{flushend}

\title{3D-MIR: A Benchmark \& Empirical Study \\on 3D Medical Image Retrieval in Radiology}
\author{
Asma Ben Abacha$^*\S$  \\
  \texttt{\small abenabacha@microsoft.com} \\
   \And
Alberto Santamaria-Pang$^*$ \\
  \texttt{\small alberto.santamariapang@microsoft.com} \\
  \And
Ho Hin Lee$^*$\\
  \texttt{\small hohinlee@microsoft.com} \\
   \AND
Jameson Merkow\\
{\tt\small jameson.merkow@microsoft.com}
 \And
Qin Cai \\
\texttt{\small Qin.Cai@microsoft.com} \\
  \And
Surya Teja Devarakonda \\
\texttt{\small sdevarakonda@microsoft.com} \\
  \And
Abdullah Islam \\
\texttt{\small Abdullah.Islam@microsoft.com} \\
  \And
Julia Gong \\
\texttt{\small juliagong@microsoft.com} \\
   \And
  Matthew P. Lungren \\
\texttt{\small mlungren@microsoft.com} \\
   \And
  Thomas Lin \\
  \texttt{\small tlin@microsoft.com} \\
     \And
  Noel C. F. Codella$^\P$ \\
  \texttt{\small ncodella@microsoft.com} \\ 
     \And
 Ivan Tarapov$^\P$ \\
 \texttt{\small Ivan.Tarapov@microsoft.com} \\
     \AND
  \texttt{Microsoft Health and Life Sciences}
}

\begin{document}
\maketitle

\def\thefootnote{\S}\footnotetext{Corresponding author.}
\def\thefootnote{*}\footnotetext{These authors contributed equally to this work and share first authorship.}
\def\thefootnote{\P}\footnotetext{ These authors contributed equally to this work and share last authorship.}

\begin{abstract}
The increasing use of medical imaging in healthcare settings presents a significant challenge due to the increasing workload for radiologists, yet it also offers opportunity for enhancing healthcare outcomes if effectively leveraged. 3D image retrieval holds potential to reduce radiologist workloads by enabling clinicians to efficiently search through diagnostically similar or otherwise relevant cases, resulting in faster and more precise diagnoses. However, the field of 3D medical image retrieval is still emerging, lacking established evaluation benchmarks, comprehensive datasets, and thorough studies. This paper attempts to bridge this gap by introducing a novel benchmark for 3D Medical Image Retrieval (3D-MIR) that encompasses four different anatomies imaged with computed tomography. Using this benchmark, we explore a diverse set of search strategies that use aggregated 2D slices, 3D volumes, and multi-modal embeddings from popular multi-modal foundation models as queries. Quantitative and qualitative assessments of each approach are provided alongside an in-depth discussion that offers insight for future research. To promote the advancement of this field, our benchmark, dataset, and code are made publicly available\footnote{\url{https://github.com/abachaa/3D-MIR}}.
\end{abstract}

% keywords can be removed

%========================
\section{Introduction} 
\label{sec:intro}
%========================

While artificial intelligence (AI) has brought great opportunities and challenges in medical imaging, its promise is tempered by the reality of increasing imaging volumes, contributing to heightened workloads and potential burnout among radiologists~\cite{harry2021physician}. Despite these challenges, the surge in medical imaging also offers a unique opportunity, by harnessing extensive databases of imaging data and clinical histories, AI can be leveraged to bolster and refine clinical decision-making, balancing the scales between the burdens and benefits of technological advancements in radiology. The key question is: \textit{How can we transform these imaging databases into a useful resource for clinicians?}
\begin{figure}[htbp]
    \centering
     \includegraphics[width=0.5\columnwidth]{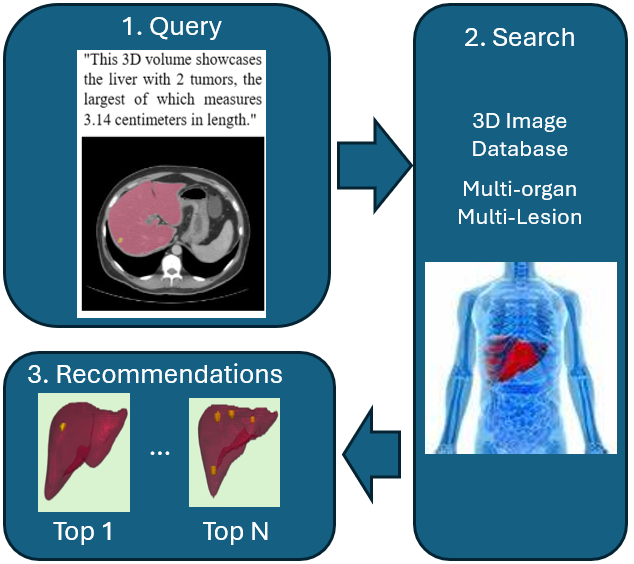}
    \caption{Schematic of the 3D Medical Image Search (3D-MIR) benchmark.}
    \label{fig:intro}
\end{figure}
To address this question, we must develop sophisticated systems to retrieve relevant and specific information from the imaging databases. Simple 2D image retrieval methods are not sufficient for complex cases that demand a more nuanced and multidimensional perspective. 3D image retrieval is promising, but it also poses many challenges, such as handling size variability and large 3D data volumes. Furthermore, 3D medical image retrieval lacks standard evaluation criteria and benchmarks, which impede research and development of advanced search methods and their clinical translation. Figure~\ref{fig:intro} illustrates an example of 3D Medical Image Retrieval (3D-MIR), specifically within CT volumes. In this example, a text description of lesion morphology in the liver is inputted, initiating a search through a database of CT scans to find liver images matching the text description. The system then presents the top \textit{n} recommendations based on this comparison.

% Lit Review
While 2D image search strategies have been extensively studied, their application to radiology is hampered by significant limitations. Traditional 2D and video image search algorithms \cite{dubey2021decade}, often fail to rank results based on medical relevance, a critical factor in clinical settings. Existing systems that utilize medical relevance are confined to 2D image searches neglecting the rich detail and complexity inherent in 3D medical imaging \cite{zhang2022contrastive}. Furthermore, even in systems that search through 3D images are limited to analyzing images slice by slice \cite{owais2019effective,qayyum2017medical} which significantly undermines the utility of 3D data. This evident gap in technology underscores the urgent need for advancements specifically designed for radiology and medical imaging workflows, where the unique challenges of 3D image retrieval can be adequately addressed.

% Impact
The potential influence of 3D search system on clinical outcomes is profound. A system that enables rapid retrieval of relevant cases has the potential to not only expedite diagnosis but also to enhance the accuracy of interpretations, thereby improving patient outcomes. 
% Objectives
This work sets out to tackle critical challenges facing advancement of 3D medical image retrieval systems.
To that end, we present a new benchmark that spans four different organ systems imaged by computed tomography (CT), specifically curated for radiology 3D image retrieval. We conduct a comparative study of various search strategies using a modern state-of-the-art medical imaging foundation model, providing both quantitative and qualitative discussion. Through these results, we shed light on the efficacy of each approach and offer valuable insights to guide future research endeavors. Our benchmark is made publicly available to facilitate further research and development on this topic.

% Closing
Our work lays the foundation for the development of systems that enable radiologists to swiftly navigate and retrieve medical 3D images, effectively turning data into a decision-support tool. This advancement streamlines the diagnostic process and fosters further research into intelligent data management in healthcare.
%========================

%========================
\section{Related Work}
%========================

3D medical imaging has predominantly concentrated on tasks such as classification~\cite{medfmc, medmnistv1, benchmd, MedMnistv2_23,RSNA_Abd23}, detection~\cite{RSNA-brain-tumor-Det2021,RSNA-spine-fracture-Det2022, RSNA_Abd23}, and segmentation~\cite{BTCV2015, DenseVnet18, MSD22,TotalSeg23, lee20223d, lee2023scaling}, while issues pertaining to diagnosis based retrieval remain under-explored. Simonyan~\textit{et al.}~\cite{simonyan2012immediate} reported the first method for 3D Region of Interest (ROI) retrieval in medical imaging. Results are demonstrated in ROI Atrophy-Aware Brain MRI Retrieval, where atlas-based model is used to compare vulnerable brain areas (e.g., hippocampal deterioration). For a comprehensive review of image retrieval systems using deep learning and hand-crafted features in medical imaging see~\cite{vishraj2022comprehensive}. 

While extensive research has been conducted in text-based image retrieval, highlighted in various studies~\cite{MedRetrv18,faiss-johnson2019,qayyum2017medical,jain2015content}, and medical image captioning\cite{pavlopoulos-etal-2019-survey, PelkaAHJFM21, RuckertAHBBISMF23, abs-2209-13983}, the integration of these techniques with the latest large-scale foundation models~\cite{Clip21, Florence21, BiomedCLIP,zhang2022contrastive} presents a significant opportunity for advancement. However, their application in 3D multi-modal medical image retrieval is still in its infancy ~\cite{lee2023region}. Current public benchmarks in 3D medical imaging, primarily focused on segmentation (e.g., MSD~\cite{MSD22}, BTCV~\cite{BTCV2015}, DenseVNet-MultiOrgan~\cite{DenseVnet18}, TCIA Pancreas-CT~\cite{pancreas_ct}, KiTS19~\cite{skm_tea}, RAD-ChestCT~\cite{rad_chestct}, ATLAS v2.0~\cite{atlas_v2}, EPISURG~\cite{episurg}) and classification (e.g., MedFMC~\cite{medfmc}, MedMNIST~\cite{medmnistv1}, BenchMD~\cite{benchmd}), do not fully address the needs of 3D image retrieval.

To address this gap, we introduce the first public benchmark, 3D-MIR, tailored for 3D multi-modal information retrieval. This benchmark includes essential elements such as full volume data, 3D labels, localized metrics at the organ level, and textual descriptions of organ lesions. Recognizing the time-intensive nature of 3D labeling and medical report generation, we leverage public datasets like the Medical Segmentation Decathlon (MSD)\cite{MSD22} for lesion segmentation, along with open-source organ segmentation models like TotalSegmentator \cite{TotalSeg23} built on \cite{NNunet21} to localize lesions in organs. We then quantify 3D lesion morphology at the organ level and utilize OpenAI's GPT foundation model~\cite{GPT4_2023} for describing organ morphology, such as the number of lesions in the liver. Our source code, data generation methods, and benchmark details are made publicly available\footnote{https://github.com/abachaa/3D-MIR}. 

%========================
\section{3D-MRI Benchmark Dataset Creation}\label{sec:benchmark} 
\begin{figure*}[h!t!p!]
    \centering
     \includegraphics[width=0.8\columnwidth]{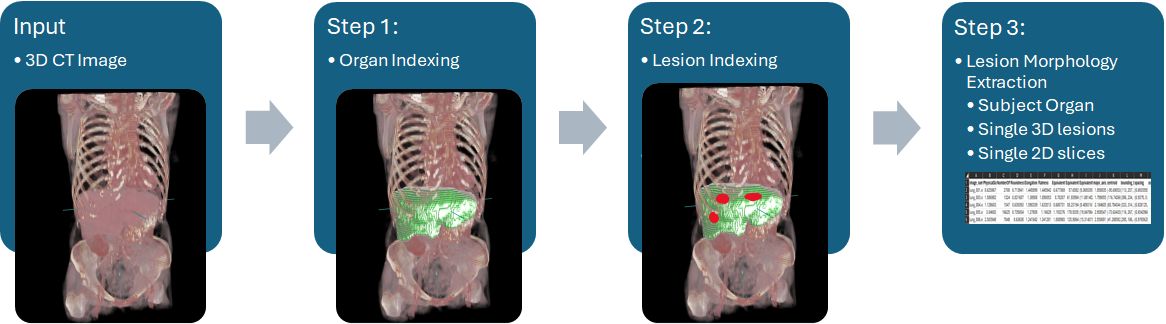}
    \caption{This schematic illustrates our 3D Medical Image Retrieval dataset benchmark, using a liver lesion example. The process begins with the full input CT volume. In Step 1, organ segmentation is applied to 104 different organs, but for this example, we only display the liver, highlighted in green. Step 2 involves identifying and indexing each individual lesion (shown in red) across the various organs. Finally, Step 3 conducts a morphological analysis of the lesions, generating specific metrics for each subject, organ, and lesion.}
    \label{fig:3dmir.pipeline}
\end{figure*}
In this section, we elaborate on the development of the 3D-MIR multi-organ dataset. Our aim is to harness public datasets that feature 3D lesion annotations to pinpoint and quantify lesions, and to provide detailed morphological descriptions of these lesions in the context of their associated organs. \textit{It is important to note that our goal diverges from producing standard clinical radiology reports or segmentations; instead, we concentrate on crafting in-depth descriptions that emphasize the morphological characteristics of the lesions to facilitate an ability to support a measure of relevance between query and search in retrieval settings.}

The 3D-MIR dataset is constructed using two primary sources: (i) the Medical Segmentation Decathlon (MSD) dataset~\cite{MSD22}, which offers 10 publicly available cohorts. Our research particularly focuses on the chest-CT cohort from the MSD, featuring 3D lesion annotations in four key organs of the chest and abdomen: the liver, colon, pancreas, and lung. For our evaluation process, we chose to work with the volumes from the training set, as the testing set volumes did not include labels. The distribution of the organs with labels in the selected volumes was as follows: liver (131 volumes), lung (64 volumes), colon (126 volumes), and a total of 281 volumes containing lesions. Figure~\ref{fig:3dmir.pipeline} shows the main steps for the dataset construction using a liver use case as example. (ii) We also utilize a comprehensive multi-organ semantic segmentation model, TotalSegmentator~\cite{TotalSeg23}, which is based on U-Net-like architectures~\cite{NNunet21,wu2018automatic}. 

\textbf{Step 1: Organ Indexing.} We begin by applying TotalSegmentator to each CT volume, which generates 104 cropped organ-CT volumes and corresponding labels for each subject~\cite{TotalSeg23}. These cropped volumes are defined by 3D bounding boxes, each encompassing the organ of interest. This initial segmentation achieves two key objectives: it localizes lesions within a comprehensive anatomical ontology and streamlines the indexing process for organ-specific queries, enhancing the efficiency and precision of data retrieval. Figure~\ref{fig:3dmir.pipeline}, Step 1 shows in green the segmented liver.

\textbf{Step 2: Lesion Indexing.} In this step, we analyze the 3D label volume associated with each CT scan to identify lesions. Given that 3D lesions are represented in a binary volume format, distinguishing lesion structures from the surrounding anatomy, we employ single-component connectivity analysis. This method effectively segregates individual lesions, allowing us to assign unique identifiers to each lesion and the corresponding subject. Subsequently, we perform quantification of each lesion's spatial overlap with the 104 organs segmented by TotalSegmentator. This quantification is crucial as it enables us to accurately map each lesion to its respective organ and lesion distribution within the organ. Such mapping is instrumental in creating a robust and precise dataset, facilitating targeted organ-specific analyses and queries.  Figure~\ref{fig:3dmir.pipeline}, Step 2 shows in red the localized individual lesion within the liver.

\textbf{Step 3: Lesion Morphology Extraction.} In this step, we focus on extracting morphological characteristics of each lesion identified in the previous steps by analyzing the 3D and 2D geometry of lesions localized in each organ. Furthermore, we extract morphological metrics corresponding to:

\textbf{a. Subject-Organ.} Subjects were grouped based on the morphological characteristics of their lesions, following the guidelines set by the American Joint Committee on Cancer's Tumor, Node, Metastasis (TNM) classification system~\cite{ajcc2017}.  This grouping adheres to the TNM system's recommendations for classifying cancer stages and provides a standardized framework for correlating lesion morphology with cancer stage. Our focus here is primarily on lesion morphology and distribution, rather than on cancer staging. Specifically, we consider the number, length, and volume of the lesions to define three groups:
\begin{itemize}
\small \setlength\itemsep{0em}
    \item[] \textbf{Lesion Group 1}: subjects with a single lesion smaller than 2 cm.
    \item[] \textbf{Lesion Group 2}  subjects with either a single lesion larger than 2 cm or multiple lesions, with none exceeding 5 cm.
    \item[] \textbf{Lesion Group 3}  subjects with one or more lesions, each larger than 5 cm.
\end{itemize}

\textbf{b. Single 3D lesion.} For each distinct lesion, we calculate comprehensive 3D morphological descriptors. These descriptors are crucial in understanding the lesion's characteristics, including: 1) Lesion topology, particularly focusing on symmetrical shapes, 2) the volume of the lesion in actual physical space, the size of the lesion, represented through the dimensions of a fitted 3D ellipsoid. For the extraction of these metrics, we have employed the ITK (Insight Segmentation and Registration Toolkit)~\cite{ITKReference}.

\textbf{c. Single 2D slices.} To complete 3D morphological analysis, we quantify 2D slices associated with lesions. In each selected slice, we measure the total area occupied by lesions, count the number of distinct lesions, and analyze their morphology, focusing on characteristics such as circularity, allowing slice-based retrieval. 

The 3D-MIR dataset is intentionally designed to facilitate the evaluation of both broad (such as distinguishing between organs with and without lesions) and detailed queries (like the number and size of lesions in each organ). It is adaptable for integration with existing public datasets that already include lesion labels. Our intention is to make this dataset publicly available in conjunction with the camera-ready version of this paper.
\begin{figure*}[h!t!p!]
\centering
     \includegraphics[width=0.9\columnwidth]{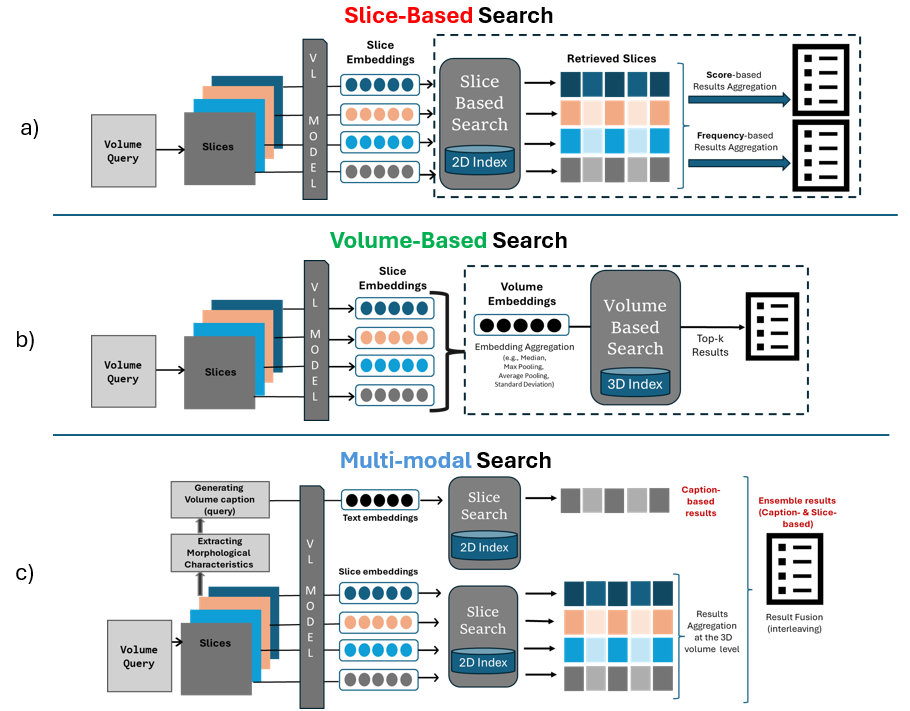}
     \caption{Overview of the retrieval methods studied. a) Slice-based. b) Volume-based. c) Multi-modal: caption-based \& ensemble.}
     \label{fig:method}
    \end{figure*}

%===========================
\section{Retrieval Methods}
\label{sec:methods}
%===========================

We investigate different search methods on several aspects. First, we study whether there is added value from the 3D perspective vs. the more common indexing and search of 2D images (in this case the 2D slices of each volume). Second, we juxtapose the performance of different approaches to evaluate the relevance and difficulty of the benchmark. We also propose and evaluate a new multi-modal ensemble method. 

In our study we rely on the BiomedCLIP vision-language model \cite{BiomedCLIP} to generate embeddings for both the 2D images and their captions. BiomedCLIP is the SOTA model on biomedical cross-modal retrieval, image classification, and visual question answering, outperforming models such as CLIP, MedCLIP and PubMedCLIP and radiology-specific SOTA models such as BioViL on radiology tasks (e.g., RSNA pneumonia detection). BiomedCLIP is pretrained on PMC-15M, a dataset of 15 million biomedical image-text pairs from over three million articles. BiomedCLIP was also improved with domain-specific adaptations for biomedical vision-language processing. The input images consist of 2D slices extracted from CT volumes, with voxel values represented in Hounsfield Units (HU). For processing, we normalize these 3D volumes from the original range of [-1000, 1000] HU to a scale of 8-bit values ranging from [0, 255]. Subsequently, from these normalized values, we generate a monochromatic image.
Our retrieval methods are described in the following sections.
\begin{figure*}[h!t!p!]
\centering
\includegraphics[width=\columnwidth]{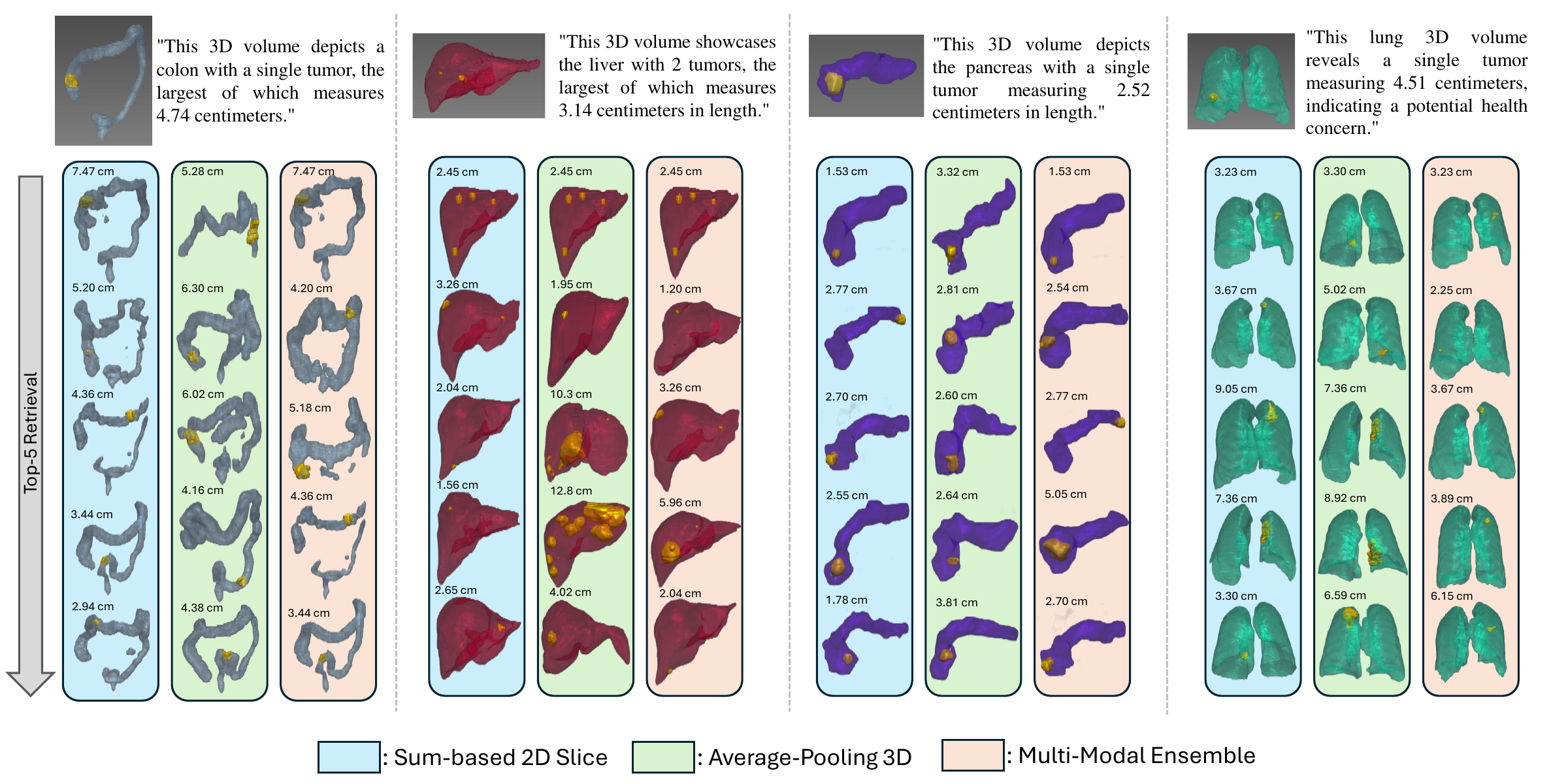}
\label{fig:sam-modalities}
 \caption{The provided overview showcases a qualitative visual representation of medical image retrieval outcomes, organized according to organ types (colon, liver, pancreas, lung) and corresponding text queries. The evaluation encompasses three methods: (i) the Sum-based 2D slice method, shown in blue to the left of each organ category, (ii) the Average Pooling-based 3D volume method, displayed in green at the center, and (iii) the Multi-modal ensemble method, featured in pink on the right. The results display the top-5 ranked outputs for each method.}
\end{figure*} %% Blue: 2D-Sum, Green: 3D-Avg, Pink: Multimodal-Ensemble 

\subsection{Slice-based Retrieval} 
\label{sec:method1}

This search method relies on the 2D slices of the 3D volumes. Each 2D slice is encoded using embeddings from the model, as shown in Figure~\ref{fig:method}a. All 2D slices of all volumes are indexed in a vector database using the Faiss library\cite{faiss-johnson2019}. 

In the search step, for a given 3D volume input $V$, the slice-based retrieval consists in using all the 2D slices of $V$ as queries (noted $Q_V$) that we use to search the vector database to retrieve the top 20 similar slices for each slice query in $Q_V$ based on the Euclidean distance. 

The retrieved 2D slices are then aggregated to their parent 3D volumes to return top-k similar images. We compared three different slice results aggregation methods based on frequency, maximum similarity score, and sum of similarity scores of the retrieved slices $R(Q_V)$ for each 3D image that has slices in the results. 

Noting the retrieved slices for a given query $Q_V$ as $R(Q_V)$, and the set of retrieved slices belonging to the same volume $V_i$ as $RS(V_i)$, the three aggregation methods rely on the following parent volume scoring methods:
\begin{equation}
\label{eqn:freq}
\text{Freq}(V) = \frac{|RS(V_i)|}{|R(Q_V)|}
\end{equation}
\begin{equation}
\label{eqn:max-score}
\text{MaxScore}(V) = \max_{S_i \in RS(V_i)} \big(\text{SimScore}(S_i)\big)
\end{equation}
\begin{equation}
\label{eqn:sum-score}
\text{ScoreSum}(V) = \sum_{S_i \in RS(V_i)} \big(\text{SimScore}(S_i)\big)
\end{equation}
In each aggregation method, the parent volumes are scored then ranked according to the respective formulas in equations \ref{eqn:freq}, \ref{eqn:max-score}, and \ref{eqn:sum-score}. The SimScore is based on the Euclidean distance between the embeddings of the slice query and the embeddings of a retrieved slice.

\subsection{Volume-based Retrieval}  
\label{sec:method2}
The volume-based retrieval relies on aggregating the embeddings of the 2D slices to generate one representative embedding vector for the whole volume, as shown in Figure~\ref{fig:method}b. 
We compared four different embedding aggregation methods: median, max pooling, average pooling, and standard deviation. Each aggregation method $Mi$ is used to generate a separate vector database/index (3D index).
In the search step, we generate the embeddings of a query volume $V$ according to $Mi$ and search for the top-k similar volumes/vectors in the corresponding 3D index. 

\subsection{Multi-modal Retrieval}
\label{sec:method3}
To create natural language queries that mimic the text that radiologists would use to query the database, we generated captions for the 3D volumes using GPT-4, OpenAI's latest GPT foundation model \cite{GPT4_2023}. The temperature parameter controls the randomness/creativity of the generated text. We selected a low temperature (temperature=0) to generate more focused/concise descriptions and to reduce hallucinations in the generated captions. Each caption describes one or more of the following image features: organ, number of lesions, and length of the largest lesion. 

Below are examples of generated captions by GPT-4 with the following main instruction in the prompt: ``\textit{You are an AI assistant with years of experience in the medical domain. 
Write a short caption that includes the following details about a 3D volume. If the number of lesion is 0, write: this is a normal image}''.
\begin{itemize}
\small \setlength\itemsep{0em}
    \item[] \textbf{Prompt}: \textit{Instructions} + Organ = liver, number of tumors = 0
    \item[] \textbf{Generated Caption:} ``A normal image of the liver with no tumors present.'' 
    \vspace{0.5em}
    \item[]  \textbf{Prompt:} \textit{Instructions} + Organ = liver, number of tumors = 11, largest tumor length in centimeter = 2.26
    \item[] \textbf{Generated Caption:} ``3D volume image showcasing a liver with 11 tumors, the largest of which measures 2.26 centimeters in length''
\end{itemize}

We used the captions to query the 2D slice index described in Section~\ref{sec:method1} through the BiomedCLIP vision-language model. The vision-language model was trained to maximize the similarly between the caption embeddings and the image embeddings using contrastive loss \cite{BiomedCLIP}.  
The results are then aggregated from the 2D slice level to the parent 3D images, based on the frequency (Equation \ref{eqn:freq}) of the retrieved slices to return top-k similar volumes. The ensemble method consists in combining search results from the caption-based method and the frequency-based slice search. Combination is based on interleaving the volumes returned by each method according to their ranks. Figure~\ref{fig:method}c describes the caption-based and ensemble multi-modal methods. 

%========================
\section{Experiments}
\label{sec:experiments}
%========================

In all our experiments, we constructed the 2D/3D index using the training slices and volumes from each dataset. To make each index collection more heterogeneous, each training set was augmented with volumes randomly selected from the other datasets. For example, the Liver training set was augmented using volumes from the Colon, Pancreas, and Lung datasets with lesions less than 2cm. This cross-reference allows to integrate `specific-healthy' organs in the evaluation dataset. The details of these training and test sets are outlined in Table~\ref{tab:data-size}. Additionally, Figure~\ref{fig:lesion.size.distribution} illustrates the size distribution of the largest lesions, measured in centimeters, across different organs, highlighting the variability in lesion size depending on the organ.

\subsection{Quantitative Evaluation}

We evaluate each search method by comparing the lesion flag and lesion group of the query volume and the top-k retrieved volumes. We then compute Precision@k (P@k) and Average Precision (AP), defined as:
\begin{equation}
AP = \sum_n (R_n - R_{n-1}) P_n
\end{equation}
\noindent\small{with $R_n$ and $P_n$ are the Precision and Recall at the nth threshold. }

%%%==============================
\begin{table}[ht]
\centering
\begin{tabular}{ |c|c|c| } 
 \hline
 \bf Organ & \bf Training Set: & \bf Test Set: \\ 
   & \#volumes / \#slices & \#volumes \\ \hline
 Liver & 157 / 58,982 & 19 \\\hline
 Colon & 209 / 32,732 & 24 \\ \hline
 Pancreas & 269 / 31,785 & 32 \\ \hline
 Lung & 94 / 23,587 & 47 \\ \hline
%  Lung - Right Upper & 82 / 21,190 & 12 \\ 
%  Lung - Left Upper & 82 / 20,734 & 12 \\ 
%  Lung - Right Lower & 82 / 19,490 & 12 \\ 
%  Lung - Left Lower & 82 / 21,197 & 12 \\ 
%  Lung - Right Middle  &  82 / 21,003 & 12 \\ \hline
\end{tabular}
\caption{Training-test splits for each dataset/organ: The training volumes/slices are used to create the 3D/2D index and the test volumes are used as test queries.}
\label{tab:data-size}
\end{table}
 
%%%==============================
\begin{figure}[htbp]
    \centering
     \includegraphics[width=\columnwidth]{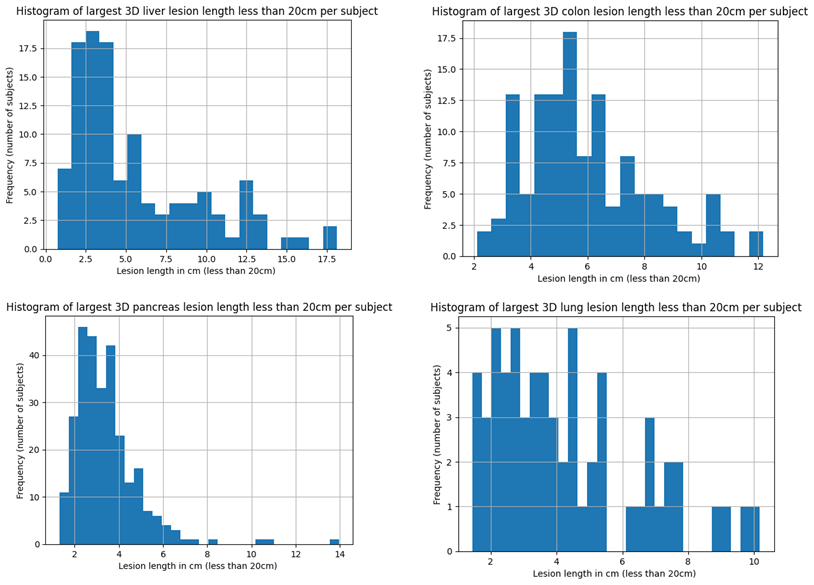}
    \caption{Lesion size distribution across organs: Liver, Colon, Pancreas, and Lung.}
    \label{fig:lesion.size.distribution}
\end{figure}

%%%==============================
%%%%%%%%% Table II %%%%%%%%%
\begin{table*}[h!]
\centering
\begin{tabular}{|p{2cm}|p{2.8cm}|p{0.9cm}|p{0.9cm}|p{0.9cm}|p{0.9cm}||p{0.9cm}|p{0.9cm}|p{0.9cm}|p{0.9cm}|} \hline
\multicolumn{2}{|l|}{\bf Average Across All Datasets}& \multicolumn{4}{|c||}{\bf Lesion Flag} & \multicolumn{4}{|c|}{\bf Lesion Group} \\\hline
\multicolumn{2}{|c|}{\bf Methods} & P@3 &  P@5 &  P@10 & AP  & P@3 &  P@5 &  P@10 & AP\\\hline
\bf Slice-based  & Frequency-based  &  86.48 &	86.17	& 85.14	& 90.55 & 61.49	& \bf 58.11	& 54.44  & \bf 69.62\\ 
\bf Search \&  & Score-based (max)  & 84.79 &	84.63	& 84.49 &	89.03 & 52.55 &	50.01	& 50.03 & 64.42\\ 
\bf Aggregation &  Score-based (sum)   & 86.48	& 85.96	& 85.12	& 90.39 & \bf 61.75	& 57.95	& \bf 54.59	& 69.49 \\  \hline
\bf Multimodal   & Generated Captions   & 58.09	& 55.54 &	55.89	& 66.19 & 37.95 &	37.47	& 37.63 & 47.92 \\ 
\bf Search & Ensemble \footnotesize{(Captions+Slices)}  & 76.78	& 74.94	& 71.86	& 82.58 & 55.09	& 53.87 &	51.90 & 65.19 \\ \hline 
\bf Volume-  & Median  &  \bf 90.22	& \bf 89.25	& \bf 87.74	& \bf 91.42 & 53.35	& 52.64	& 52.72 & 64.61 \\ 
\bf  based & Max Pooling  & 39.51 &	40.48	& 48.52 & 52.75	& 27.93 & 27.74 &	30.73 & 39.61 \\  
\bf Search & Average Pooling & 88.29 & 86.96	& 86.46 & 90.43	& 59.10	& 54.79	& 53.03  & 67.33 \\ 
& Standard deviation  & 53.74	& 59.33 &	58.02 & 61.40	& 32.05 &	31.65 &	29.22 & 39.82 \\  \hline
\end{tabular}
\caption{Overall Average Results across all Datasets}
\label{tab:avg-results}
\end{table*}
%
 
%%%==============================
%%%%%%%%% Table I (Liver) %%%%%%%%%
\begin{table*}[h!]
\centering
\begin{tabular}{|p{2cm}|p{2.8cm}|p{0.9cm}|p{0.9cm}|p{0.9cm}|p{0.9cm}||p{0.9cm}|p{0.9cm}|p{0.9cm}|p{0.9cm}|} \hline
\multicolumn{2}{|l|}{\bf Liver Dataset}& \multicolumn{4}{|c||}{\bf Lesion Flag} & \multicolumn{4}{|c|}{\bf Lesion Group} \\\hline
\multicolumn{2}{|c|}{\bf Methods} & P@3 &  P@5 &  P@10 & AP & P@3 &  P@5 &  P@10 & AP \\\hline
\bf Slice-based  & Frequency-based  & 82.46   & \bf 81.05  & \bf 78.18 & \bf 87.98 & \bf 57.89 & \bf 51.58  &  \bf 49.19 & \bf 64.37 \\ 
\bf Search \& & Score-based (max)  & 71.93 & 74.74 & 76.65 & 79.93 &  43.86 & 41.05 & 42.01 & 55.08 \\ 
\bf Aggregation &  Score-based (sum)   &  82.46 & \bf 81.05 & \bf 78.18 & 87.54 & \bf 57.89 &  \bf 51.58 & \bf 49.19 & 63.93 \\\hline
\bf Multimodal   & Generated Captions & 70.18 & 61.05 & 61.05 & 77.95 & 36.84 & 34.74 & 34.74 & 53.63 \\
\bf Search & Ensemble \footnotesize{(Captions+Slices)} & \bf 84.21 & \bf 81.05 & 72.44 & 83.72 & 50.88 & 50.53 & 43.83 & 59.32 \\\hline 
\bf Volume-  &  Median & 78.95 & 77.89 & 76.94 & 81.57 & 40.35 & 43.16 & 44.59 & 50.20 \\
\bf based & Max Pooling & 68.42 & 61.05 & 65.36 & 69.66 & 22.81 & 25.26 & 29.09 & 40.65 \\
\bf Search & Average Pooling  &  73.68  &  74.74 &  74.74 & 80.22 &  47.37   &  48.42 &  45.55 & 55.58 \\
& Standard Deviation &  68.42  &  68.42 &  68.42 & 68.42 & 40.35   &  34.74 &  32.34 & 41.23 \\ 
 \hline
\end{tabular}
\caption{Precision@k and Average Precision on the Liver Dataset}
\label{tab:results-liver}
\end{table*}
 
%%%==============================
%%%%%%%%% Table II %%%%%%%%%
\begin{table*}[h!]
\centering
\begin{tabular}{|p{2cm}|p{2.8cm}|p{0.9cm}|p{0.9cm}|p{0.9cm}|p{0.9cm}||p{0.9cm}|p{0.9cm}|p{0.9cm}|p{0.9cm}|} \hline
\multicolumn{2}{|l|}{\bf Colon Dataset}& \multicolumn{4}{|c||}{\bf Lesion Flag} & \multicolumn{4}{|c|}{\bf Lesion Group} \\\hline
\multicolumn{2}{|c|}{\bf Methods} & P@3 &  P@5 &  P@10 & AP & P@3 &  P@5 &  P@10 & AP\\\hline
\bf Slice-based  & Frequency-based  &  70.83  &  74.17 & 73.79 & 79.55 &  55.56  &  \bf 54.17 &  51.52 & 62.66 \\ 
\bf Search \&  & Score-based (max) & 77.78 & 76.67 & 75.53 & 86.00 & 50.00 & 46.67 & 45.53 & 61.49  \\ 
\bf Aggregation &  Score-based (sum)   & 70.83 & 73.33 & 73.71 & 79.30 & 55.56 & \bf 54.17 & \bf 51.89 & 62.44 \\\hline 
\bf Multimodal   & Generated Captions  & 45.83 & 45.83 & 45.83 & 60.51 & 36.11 & 36.11 & 36.11 & 45.31\\ 
\bf Search & Ensemble \footnotesize{(Captions+Slices)}  & 65.28 & 61.67 & 59.40 & 69.80 &  55.56 & 49.17 & 46.31 & 56.16 \\\hline 
\bf Volume-  & Median & \bf 84.72 & \bf 83.33 & \bf 78.79 & \bf 87.09 & 51.39 & 51.67 & 47.88 & 63.30 \\ 
\bf  based & Max Pooling & 41.67 & 46.67 & 48.56 & 58.32 & 41.67 & 40.00 & 34.70 & 49.62 \\
\bf Search & Average Pooling  & \bf 84.72  &  78.33 & 77.58 & 85.90 & \bf 56.94  &  50.00 & 48.86 & \bf 67.13 \\ 
& Standard deviation & 58.33  &  60.83 &  57.05 & 63.66 &  23.61  &  25.00 &  25.00 & 35.69 \\ 
 \hline
\end{tabular}
\caption{Precision@k and Average Precision on the Colon Dataset}
\label{tab:results-colon}
\end{table*} 
%%%==============================
%%%%%%%%% Table III %%%%%%%%%
\begin{table*}[h!]
\centering
\begin{tabular}{|p{2cm}|p{2.8cm}|p{0.9cm}|p{0.9cm}|p{0.9cm}|p{0.9cm}||p{0.9cm}|p{0.9cm}|p{0.9cm}|p{0.9cm}|} \hline
\multicolumn{2}{|l|}{\bf Pancreas Dataset} & \multicolumn{4}{|c||}{\bf LesionFlag} & \multicolumn{4}{|c|}{\bf Lesion Group} \\\hline
\multicolumn{2}{|c|}{\bf Methods} & P@3 &  P@5 &  P@10 & AP & P@3 &  P@5 &  P@10 & AP \\\hline
\bf Slice-based  & Frequency-based  &  96.88   & 95.00 & 95.85 & 98.81 & 55.21 &  \bf 55.63  & 54.49 & 65.78 \\ 
\bf Search \&  & Score-based (max)  & 95.83 & 95.63 & 94.49 & 95.69 & 48.96 & 51.87 & 52.73 & 63.23 \\ 
\bf Aggregation & Score-based (sum)  & 96.88 & 95.00 & 95.85 & \bf 98.88 & 56.25 & 55.00 & 54.72 & \bf 66.14 \\\hline
\bf Multimodal   & Generated Captions  & 82.29 & 81.25 & 82.64 & 83.30 & 44.79 & 45.00 & 45.62 & 57.11 \\ 
\bf Search & Ensemble \footnotesize{(Captions+Slices)} & 90.62 & 90.00 & 88.58 & 92.70 & 46.88 & 48.75 & 50.45 & 58.66 \\\hline 
\bf Volume- & Median & \bf 97.92 & \bf 97.50 & \bf 96.93 & 97.94 & 47.92 & 50.63 & 56.59 & 61.90 \\
\bf based   & Max pooling & 12.50 & 12.50 & 32.67 & 30.15 & 12.50 & 12.50 & 25.57 & 24.52  \\ 
\bf Search & Average Pooling  &  96.88  & 96.88 &  96.59 & 97.06 & \bf 58.33   & 55.62 &  \bf 57.61 & 63.25 \\
& Standard deviation &  54.17   &  61.25 &  64.66 & 64.17 & 30.21   & 26.88 &  25.17 & 39.71 \\\hline
\end{tabular}
\caption{Precision@k and Average Precision on the Pancreas Dataset}
\label{tab:results-pancreas}
\end{table*}

%%%==============================
%%%%%%%%% Table I (Liver) %%%%%%%%%
\begin{table*}[h!]
\centering
\begin{tabular}{|p{2cm}|p{2.8cm}|p{0.9cm}|p{0.9cm}|p{0.9cm}|p{0.9cm}||p{0.9cm}|p{0.9cm}|p{0.9cm}|p{0.9cm}|} \hline
\multicolumn{2}{|l|}{\bf Lung Dataset}& \multicolumn{4}{|c||}{\bf Lesion Flag} & \multicolumn{4}{|c|}{\bf Lesion Group} \\\hline
\multicolumn{2}{|c|}{\bf Methods} & P@3 &  P@5 &  P@10  & AP & P@3 &  P@5 &  P@10 & AP \\\hline
\bf Slice-based  & Frequency-based  & 95.74 & 94.47 & 92.73 & 95.86 & \bf 77.30 & \bf 71.06 &  62.55 & 85.69 \\ 
\bf Search  \& & Score-based (max)  & 93.62  & 91.49 & 91.30 & 94.49 & 67.38 & 60.43 &  59.85 & 77.88 \\ 
\bf Aggregation &  Score-based (sum)  & 95.74  & 94.47 & 92.73 & 95.86 & \bf 77.30 & \bf 71.06 &  62.55 & 85.46 \\ \hline
\bf Multimodal   & Generated Captions & 34.04  & 34.04 & 34.04 & 43.00 & 34.04 & 34.04 &  34.04 & 35.65 \\ 
\bf Search & Ensemble \footnotesize{(Captions+Slices)} & 67.02  & 67.02 & 67.02 & 84.11 & 67.02 & 67.02 & \bf 67.02 & \bf 86.61\\ \hline 
\bf Volume-  & Median & \bf 99.29  & \bf 98.30 & \bf 98.30 & \bf 99.07 & 73.76 & 65.11 & 61.82 & 83.06 \\ 
\bf based & Max Pooling & 35.46  & 41.70 & 47.50 & 52.89 & 34.75 & 33.19 &  33.58 & 43.64 \\ 
\bf Search & Average Pooling  & 97.87  & 97.87 & 96.91 & 98.54 & 73.76 & 65.11 &  60.08 & 83.35 \\ 
& Standard Deviation   & 34.04 & 46.81 & 41.97 & 49.36 & 34.04 & 40.00 &  34.39 & 42.67 \\ \hline
\end{tabular}
\caption{Precision@k and Average Precision on the Lung Dataset}
\label{tab:results-lung}
\end{table*}
 
%%%==============================

The overall results are presented in Table~\ref{tab:avg-results} and the specific organ-wise results are listed in Table~\ref{tab:results-liver}, Table~\ref{tab:results-colon}, Table~\ref{tab:results-pancreas}, and Table~\ref{tab:results-lung} for the Liver, Colon, Pancreas, and Lung, respectively.

In three out of four datasets, the \textbf{lesion flag} is best recognized by the volume-based search method with median pooling used to define the volume embeddings. This demonstrates the benefits of using the 3D context to accurately detect lesions. Given that ground truth data was generated using a classification model, the substantially better performance of the volume-based search method was likely due to reduced the signal noise over the individual 2D slice embeddings. 

In three out of the four datasets, the \textbf{lesion group} is most effectively identified by either the slice-based search method or the multi-modal search method. This is likely because the lesion group frequently correlates to its size relative to the organ (lesion size distribution is shown in Figure~\ref{fig:lesion.size.distribution}). A cumulative perspective (such as the sum) of the data in the individual slices appears to better represent the lesion's significance within the organs. In contrast to a volume-based representation which aggregates embeddings from all the slices into a single global semantic representation, potentially losing the detailed cumulative information about the lesion's relative size. 

The multi-modal ensemble method, combining the results of the caption-based and the slice-based methods, achieved the best precision@3 of 84.21\% for lesion flag matching on the Liver dataset and the best precision@10 of 67.02\% and average precision of 86.61\% for lesion group matching on the Lung dataset. However, overall, this ensemble with results fusion underperformed compared to the slice-based and volume-based methods. This is likely due to the relatively low similarity between the caption embeddings and the image embeddings. 

%%%%%%%%%%%%%%%%%%%%%%%%%%% 

\subsection{Discussion}
In our analysis, we not only measure quantitative aspects but also investigate how well different methods correlate the retrieval of organs with their lesions by creating a 3D surface visualization of lesion groups for each approach. Initially focusing on the liver, we find that both 2D slice-based and multi-modal methods show adeptness in identifying smaller lesions located near the liver's periphery. In contrast, the 3D volumetric approach excels in retrieving images of larger lesions, despite showing a lower correlation between the query and the results.

When searching for colon lesions, the results are consistent across all methods, with each showing a strong correlation in terms of accurately determining the lesion group. However, we observe a variation in the accuracy of lesion location identification. The 2D slice-based method is particularly effective in locating lesions in the left section of the colon.

Regarding the pancreas, foundation model representations are notably successful in identifying lesion regions, especially at the lower end of the organ. Despite some limitations in classifying the group-size of pancreatic lesions (cf. Table \ref{tab:results-pancreas}), the 3D approach shows adaptability in handling complex lesion morphologies like U-shape configurations.

From our preliminary analysis of the 2D slice-based embedding, it appears that tasks involving broader categorization, such as lesion detection (lesion flag), are more effective than those requiring more detailed analysis, like lesion. Additionally, our findings indicate that no single method excels uniformly across all types of organs. This observation suggests that the variability between different organs significantly influences the effectiveness of encoding both lesion information and the appearance of the organs. In our initial experiments, we examined the use of 2D embedding techniques for complex 3D tasks. The performance issues we encountered might be linked to pre-analytical variables like image resolution and normalization, as well as to biological and anatomical differences, such as varying organ sizes and distinct organ characteristics. Despite applying 2D embeddings in scenarios beyond their typical domain, they proved to be useful, particularly in identifying specific morphological features and patterns in 3D data. This indicates that, although 2D embeddings may not completely capture the nuances of 3D tasks, they can still significantly enhance analysis, especially when integrated with other dimensional techniques. 

\section{Conclusion} 
\label{sec:conclusion}
This paper introduces the first benchmark for 3D Medical Image Retrieval (3D-MIR), covering four types of anatomies imaged via computed tomography, and evaluates a range of search strategies leveraging popular state-of-art multi-modal foundation models. These strategies involve queries based on aggregated 2D slices, 3D volumes, or multi-modal embeddings. Our findings indicate that while current foundation models are adept at identifying coarse-grained semantic details, such as lesion presence, they struggle with fine-grained information, such as grouping of lesions by size. Given that the existing foundation models are designed for 2D inputs, our research concentrates on investigating two approaches to combine 2D information into a 3D context for 3D image search: slice-based and volume-based. Our experimental outcomes hint that volume-based approaches may be more effective for broad categorizations, whereas slice-based methods could be a more effective choice to capturing fine-grained details. Finally, our results demonstrate that there remains significant margin for further performance improvement across all tasks. This benchmark is made publicly available to support continued progress of methods and foundation models for 3D image retrieval.

\section*{Acknowledgements}
We thank Paul Vozila for his valuable comments and discussion on the paper.

%%%%%%%%% REFERENCES %%%%%%%%%
\bibliographystyle{unsrt}
\bibliography{references}

\begin{thebibliography}{10}

\bibitem{harry2021physician}
Elizabeth Harry, Christine Sinsky, Lotte~N Dyrbye, Maryam~S Makowski, Mickey Trockel, Michael Tutty, Lindsey~E Carlasare, Colin~P West, and Tait~D Shanafelt.
\newblock Physician task load and the risk of burnout among us physicians in a national survey.
\newblock {\em The Joint Commission Journal on Quality and Patient Safety}, 47(2):76--85, 2021.

\bibitem{dubey2021decade}
Shiv~Ram Dubey.
\newblock A decade survey of content based image retrieval using deep learning.
\newblock {\em IEEE Transactions on Circuits and Systems for Video Technology}, 32(5):2687--2704, 2021.

\bibitem{zhang2022contrastive}
Yuhao Zhang, Hang Jiang, Yasuhide Miura, Christopher~D Manning, and Curtis~P Langlotz.
\newblock Contrastive learning of medical visual representations from paired images and text.
\newblock In {\em Machine Learning for Healthcare Conference}, pages 2--25. PMLR, 2022.

\bibitem{owais2019effective}
Muhammad Owais, Muhammad Arsalan, Jiho Choi, and Kang~Ryoung Park.
\newblock Effective diagnosis and treatment through content-based medical image retrieval (cbmir) by using artificial intelligence.
\newblock {\em Journal of clinical medicine}, 8(4):462, 2019.

\bibitem{qayyum2017medical}
Adnan Qayyum, Syed~Muhammad Anwar, Muhammad Awais, and Muhammad Majid.
\newblock Medical image retrieval using deep convolutional neural network.
\newblock {\em Neurocomputing}, 266:8--20, 2017.

\bibitem{medfmc}
Dequan Wang, Xiaosong Wang, Lilong Wang, Mengzhang Li, Qian Da, Xiaoqiang Liu, Xiangyu Gao, Jun Shen, Junjun He, Tian Shen, et~al.
\newblock Medfmc: A real-world dataset and benchmark for foundation model adaptation in medical image classification.
\newblock {\em arXiv preprint arXiv:2306.09579}, 2023.

\bibitem{medmnistv1}
Jiancheng Yang, Rui Shi, and Bingbing Ni.
\newblock Medmnist classification decathlon: A lightweight automl benchmark for medical image analysis.
\newblock In {\em IEEE 18th International Symposium on Biomedical Imaging (ISBI)}, pages 191--195, 2021.

\bibitem{benchmd}
Kathryn Wantlin, Chenwei Wu, Shih{-}Cheng Huang, Oishi Banerjee, Farah Dadabhoy, Veeral~Vipin Mehta, Ryan~Wonhee Han, Fang Cao, Raja~R. Narayan, Errol Colak, Adewole~S. Adamson, Laura Heacock, Geoffrey~H. Tison, Alex Tamkin, and Pranav Rajpurkar.
\newblock Benchmd: {A} benchmark for modality-agnostic learning on medical images and sensors.
\newblock {\em CoRR}, abs/2304.08486, 2023.

\bibitem{MedMnistv2_23}
Jiancheng Yang, Rui~Shi ad~Donglai~Wei, Zequan Liu, Lin Zhao, Bilian Ke, Hanspeter Pfister, and Bingbing Ni.
\newblock Medmnist v2 - a large-scale lightweight benchmark for 2d and 3d biomedical image classification.
\newblock {\em Scientific Data}, 10(41), 2023.

\bibitem{RSNA_Abd23}
Radiological~Society of~North~America.
\newblock Rsna abdominal trauma detection ai challenge, 2023.

\bibitem{RSNA-brain-tumor-Det2021}
Rsna brain tumor ai challenge.
\newblock 2021.

\bibitem{RSNA-spine-fracture-Det2022}
Rsna cervical spine fracture ai challenge.
\newblock 2022.

\bibitem{BTCV2015}
Miccai 2015 multi-atlas abdomen labeling challenge.
\newblock 2015.

\bibitem{DenseVnet18}
E.~Gibson, F.~Giganti, Y.~Hu, E.~Bonmati, S.~Bandula, K.~Gurusamy, B.~Davidson, S.~P. Pereira, M.~J. Clarkson, and D.~C. Barratt.
\newblock Automatic multi-organ segmentation on abdominal ct with dense v-networks.
\newblock {\em IEEE Trans Med Imaging}, 37(8):1822–1834, 2018.

\bibitem{MSD22}
M.~Antonelli et~al.
\newblock The medical segmentation decathlon.
\newblock {\em Nature Communications}, 13(4128), 2022.

\bibitem{TotalSeg23}
Jakob Wasserthal, Hanns-Christian Breit, Manfred~T. Meyer, Maurice Pradella, Daniel Hinck, Alexander~W. Sauter, Tobias Heye, Daniel~T. Boll, Joshy Cyriac, Michael~Bach Shan~Yang, and Martin Segeroth.
\newblock Totalsegmentator: Robust segmentation of 104 anatomic structures in ct images.
\newblock {\em Radiology: Artificial Intelligence}, 5(5), 2023.

\bibitem{lee20223d}
Ho~Hin Lee, Shunxing Bao, Yuankai Huo, and Bennett~A Landman.
\newblock 3d ux-net: A large kernel volumetric convnet modernizing hierarchical transformer for medical image segmentation.
\newblock {\em arXiv preprint arXiv:2209.15076}, 2022.

\bibitem{lee2023scaling}
Ho~Hin Lee, Quan Liu, Shunxing Bao, Qi~Yang, Xin Yu, Leon~Y Cai, Thomas~Z Li, Yuankai Huo, Xenofon Koutsoukos, and Bennett~A Landman.
\newblock Scaling up 3d kernels with bayesian frequency re-parameterization for medical image segmentation.
\newblock In {\em International Conference on Medical Image Computing and Computer-Assisted Intervention}, pages 632--641. Springer, 2023.

\bibitem{simonyan2012immediate}
K.~Simonyan, M.~Modat, S.~Ourselin, A.~Criminisi, A.~Zisserman, and Antonio Criminisi.
\newblock Immediate roi search for 3d medical images.
\newblock In {\em MICCAI workshop on Medical Content-based Retrieval for Clinical Decision Support (MCBR-CDS)}, October 2012.

\bibitem{vishraj2022comprehensive}
Rashmi Vishraj, Savita Gupta, and Sukhwinder Singh.
\newblock A comprehensive review of content-based image retrieval systems using deep learning and hand-crafted features in medical imaging: Research challenges and future directions.
\newblock {\em Computers and Electrical Engineering}, 104:108450, 2022.

\bibitem{MedRetrv18}
Z.~Li, X.~Zhang, H.~Müller, and S.~Zhang.
\newblock Large-scale retrieval for medical image analytics: A comprehensive review.
\newblock {\em Med Image Anal.}, (43):66--84, 2018.

\bibitem{faiss-johnson2019}
Jeff Johnson, Matthijs Douze, and Herv{\'e} J{\'e}gou.
\newblock Billion-scale similarity search with {GPUs}.
\newblock {\em IEEE Transactions on Big Data}, 7(3):535--547, 2019.

\bibitem{jain2015content}
Sahil Jain, Kiranmai Pulaparthi, and Chetan Fulara.
\newblock Content based image retrieval.
\newblock {\em Int. J. Adv. Eng. Glob. Technol}, 3:1251--1258, 2015.

\bibitem{pavlopoulos-etal-2019-survey}
John Pavlopoulos, Vasiliki Kougia, and Ion Androutsopoulos.
\newblock A survey on biomedical image captioning.
\newblock In Raffaella Bernardi, Raquel Fernandez, Spandana Gella, Kushal Kafle, Christopher Kanan, Stefan Lee, and Moin Nabi, editors, {\em Proceedings of the Second Workshop on Shortcomings in Vision and Language}, pages 26--36, Minneapolis, Minnesota, June 2019. Association for Computational Linguistics.

\bibitem{PelkaAHJFM21}
Obioma Pelka, Asma~Ben Abacha, Alba Garcia~Seco de~Herrera, Janadhip Jacutprakart, Christoph~M. Friedrich, and Henning M{\"{u}}ller.
\newblock Overview of the imageclefmed 2021 concept {\&} caption prediction task.
\newblock In Guglielmo Faggioli, Nicola Ferro, Alexis Joly, Maria Maistro, and Florina Piroi, editors, {\em Proceedings of the Working Notes of {CLEF} 2021 - Conference and Labs of the Evaluation Forum, Bucharest, Romania, September 21st - to - 24th, 2021}, volume 2936 of {\em {CEUR} Workshop Proceedings}, pages 1101--1112. CEUR-WS.org, 2021.

\bibitem{RuckertAHBBISMF23}
Johannes R{\"{u}}ckert, Asma~Ben Abacha, Alba Garcia~Seco de~Herrera, Louise Bloch, Raphael Br{\"{u}}ngel, Ahmad Idrissi{-}Yaghir, Henning Sch{\"{a}}fer, Henning M{\"{u}}ller, and Christoph~M. Friedrich.
\newblock Overview of imageclefmedical 2023 - caption prediction and concept detection.
\newblock In Mohammad Aliannejadi, Guglielmo Faggioli, Nicola Ferro, and Michalis Vlachos, editors, {\em Working Notes of the Conference and Labs of the Evaluation Forum {(CLEF} 2023), Thessaloniki, Greece, September 18th to 21st, 2023}, volume 3497 of {\em {CEUR} Workshop Proceedings}, pages 1328--1346. CEUR-WS.org, 2023.

\bibitem{abs-2209-13983}
Alexander Selivanov, Oleg~Y. Rogov, Daniil Chesakov, Artem Shelmanov, Irina Fedulova, and Dmitry~V. Dylov.
\newblock Medical image captioning via generative pretrained transformers.
\newblock {\em Scientific Reports}, 13, 2023.

\bibitem{Clip21}
A.~Radford, J.~W. Kim, C.~Hallacy, A.~Ramesh, G.~Goh, S.~Agarwa, G.~Sastry, Askell, P.~Mishkin, J.~Clark, G.~Krueger, and I.~Sutskever.
\newblock Learning transferable visual models from natural language supervision.
\newblock {\em arXiv 2103.00020}, 2021.

\bibitem{Florence21}
Y.~Lu et. al.
\newblock Florence: A new foundation model for computer vision.
\newblock {\em arXiv 2111.11432v1}, 2021.

\bibitem{BiomedCLIP}
Sheng Zhang, Yanbo Xu, Naoto Usuyama, Jaspreet Bagga, Robert Tinn, Sam Preston, Rajesh Rao, Mu~Wei, Naveen Valluri, Cliff Wong, Matthew~P. Lungren, Tristan Naumann, and Hoifung Poon.
\newblock Large-scale domain-specific pretraining for biomedical vision-language processing.
\newblock {\em CoRR}, abs/2303.00915, 2023.

\bibitem{lee2023region}
Ho~Hin Lee, Alberto Santamaria-Pang, Jameson Merkow, Ozan Oktay, Fernando P{\'e}rez-Garc{\'\i}a, Javier Alvarez-Valle, and Ivan Tarapov.
\newblock Region-based contrastive pretraining for medical image retrieval with anatomic query.
\newblock {\em arXiv preprint arXiv:2305.05598}, 2023.

\bibitem{pancreas_ct}
Sean Berryman.
\newblock Pancreas-ct.
\newblock {\em The Cancer Imaging Archive (TCIA) Public Access}, 2023.

\bibitem{skm_tea}
StanfordMIMI.
\newblock Stanford knee mri multi-task evaluation (skm-tea) dataset.
\newblock {\em GitHub}, 2023.

\bibitem{rad_chestct}
Kenneth Clark, Bruce Vendt, Kirk Smith, John Freymann, Justin Kirby, Paul Koppel, Stephen Moore, Stanley Phillips, David Maffitt, Michael Pringle, et~al.
\newblock Rad-chestct dataset.
\newblock {\em CVIT - Center for Virtual Imaging Trials}, 2023.

\bibitem{atlas_v2}
Yansong Ma, Yiming Li, Xiaoying Li, Xia Li, Shuo Li, Jie Zhang, Yilong Zhang, Yongjun Wang, Xinfeng Liu, David Wang, et~al.
\newblock A large, curated, open-source stroke neuroimaging dataset to advance machine learning in clinical research.
\newblock {\em Nature Scientific Data}, 9(1):1--10, 2022.

\bibitem{episurg}
Shuyue Wang, Zhe Wang, Jingwei Zhang, Jian Zhang, Zhiqiang Zhang, Shuang Wang, Xiaojun Wang, Yujie Wang, Shuyu Li, Yinyan Wang, et~al.
\newblock Episurg: An epilepsy surgery mri dataset.
\newblock {\em Data in brief}, 32:106231, 2020.

\bibitem{NNunet21}
F.~Isensee, P.~F. Jaeger, S.~A. Kohl, J.~Petersen, and K.~H. Maier-Hein.
\newblock nnu-net: a self-configuring method for deep learning-based biomedical image segmentation.
\newblock {\em Nature methods}, 18(2), 2021.

\bibitem{GPT4_2023}
OpenAI.
\newblock Gpt-4 technical report.
\newblock {\em ArXiv}, abs/2303.08774, 2023.

\bibitem{wu2018automatic}
Xing Wu, Xiangrong Zhou, Qiang Zhang, Jiawei Tian, Shuo Wang, Lei Li, Xiaoyun Zhang, Dongni Wang, Xiaojie Wang, Xiaosong Wang, et~al.
\newblock Automatic multi-organ segmentation on abdominal ct with dense v-networks.
\newblock {\em IEEE Transactions on Medical Imaging}, 37(8):1822--1834, 2018.

\bibitem{ajcc2017}
Mahul~B Amin, Stephen~B Edge, Frederick~L Greene, David~R Byrd, Rebecca~K Brookland, Mary~Kay Washington, Jeffrey~E Gershenwald, Carolyn~C Compton, Kenneth~R Hess, Daniel~C Sullivan, et~al.
\newblock {\em AJCC Cancer Staging Manual}.
\newblock Springer, 2017.

\bibitem{ITKReference}
Hans~J. Johnson, Matthew~M. McCormick, Luis Ibanez, and The~ITK Collaboration.
\newblock The insight segmentation and registration toolkit.
\newblock {\em Software Guide, Updated for ITK version 5.2.0}, 2021.

\end{thebibliography}

\end{document}